\documentclass{article}

\usepackage[preprint]{neurips_wrl2020} 

\usepackage[utf8]{inputenc} 
\usepackage[T1]{fontenc}    
\usepackage{url}            
\usepackage{booktabs}       
\usepackage{amsfonts}       
\usepackage{nicefrac}       
\usepackage{microtype}      
\usepackage{microtype}
\usepackage[dvips]{graphicx}
\usepackage{xcolor}
\usepackage{times}
\usepackage{float}
\usepackage{amsmath}
\usepackage{amssymb}
\usepackage{mathabx}
\usepackage{dsfont}
\usepackage[ruled,vlined,noresetcount]{algorithm2e}
\usepackage{cancel}
\usepackage{multirow}
\usepackage{subcaption}
\usepackage{blindtext}
\usepackage{amssymb}
\usepackage{wrapfig}
\title{Model-based Policy Search\\ for Partially Measurable Systems}

\newcommand*\samethanks[1][\value{footnote}]{\footnotemark[#1]}

\author{%
  Fabio Amadio\thanks{Deptartment of Information Engineering,  University of Padova, Via Gradenigo 6/B, 35131 Padova, Italy}\\
  \texttt{amadiofa@dei.unipd.it} \\
  \And
  Alberto Dalla Libera\samethanks \\
  \texttt{dallaliber@dei.unipd.it} \\
  \And 
  Ruggero Carli\samethanks  \\
  \texttt{carlirug@dei.unipd.it} \\
  \And
  Daniel Nikovski\thanks{Mitsubishi Electric Research Laboratories (MERL), Cambridge, MA 02139}\\
  \texttt{nikovski@merl.com} \\
  \And
  Diego Romeres\samethanks\\
  \texttt{romeres@merl.com} \\
}

\begin{document}

\maketitle

\begin{abstract}
In this paper, we propose a Model-Based Reinforcement Learning (MBRL) algorithm for Partially Measurable Systems (PMS), i.e., systems where the state can not be directly measured, but must be estimated through proper state observers. The proposed algorithm,  named Monte Carlo Probabilistic Inference for Learning COntrol for Partially Measurable Systems (MC-PILCO4PMS), relies on Gaussian Processes (GPs) to model the system dynamics, and on a Monte Carlo approach to update the policy parameters. W.r.t. previous GP-based MBRL algorithms, MC-PILCO4PMS models explicitly the presence of state observers during policy optimization, allowing to deal PMS. The effectiveness of the proposed algorithm has been tested both in simulation and in two real systems.
\end{abstract}

\section{Introduction}\label{sec:introduction}
Reinforcement Learning (RL) \cite{sutton2018reinforcement} has achieved outstanding results in many different environments. MBRL algorithms seem a promising solution to apply RL to real systems, due to their data-efficiency w.r.t. model-free RL algorithms. In particular, remarkable results have been obtained relying on Gaussian Processes (GPs) \cite{williams2006gaussian} to model the systems dynamics, see for instance \cite{deisenroth2011pilco,parmas2018pipps,chatzilygeroudis2017black,romeres2019semiparametrical,romeres2019derivative}. In this paper, we cosider the application of MBRL algorithms to PMS, i.e., systems where only a subset of the state components can be directly measured, and the remaining components can be estimated through proper state observer. PMS are particularly relevant in real world applications, think for example to mechanical systems, where, typically, only positions are measured, while velocities are estimated thorough numerical differentiation or more complex filters. The proposed algorithm, named MC-PILCO4PMS, relies on Gaussian Processes (GPs) to model the system dynamics, and on a Monte Carlo approach \cite{caflisch1998monte} to optimize the policy parameters. W.r.t. previous GP-based MBRL algorithms, such as  \cite{deisenroth2011pilco,parmas2018pipps,chatzilygeroudis2017black,dalla2020model}, MC-PILCO4PMS models explicitly the presence of two different state observers during the two phases of model learning and of policy optimization. This improves the characterization of the PMS in the two phases and so the control performance. In the following we provide a description of the proposed algorithm, assuming that it is applied to mechanical systems where only positions measurement are available. However, the algorithm generalizes to any PMS. 

\section{Problem Setting}
Consider a mechanical system with $d_{\boldsymbol{q}}$ degrees of freedom, and denote with $\boldsymbol{x}_{t} = [\boldsymbol{q}_{t}^T,\boldsymbol{\dot{q}}_{t}^T]^T$ its state, where $\boldsymbol{q}_{t} \in \mathbb{R}^{d_{\boldsymbol{q}}}$ and $\boldsymbol{\dot{q}}_{t}\in \mathbb{R}^{d_{\boldsymbol{q}}}$ are, respectively, the vector of the generalized coordinates and its derivative w.r.t. time. Assume that joint positions can be directly measured, while $\boldsymbol{\dot{q}}_t$ must be estimated from the history of $\boldsymbol{q}_t$ measurements. Moreover, let the system be Markovian, and describe its discrete-time dynamics as $\boldsymbol{x}_{t+1} = f(\boldsymbol{x}_{t}, \boldsymbol{u}_{t}) + \boldsymbol{w}_{t}$, where $f(\cdot)$ is an unknown transition function, $\boldsymbol{u}_{t} \in \mathbb{R}^{d_{\boldsymbol{u}}}$ represents the system input, while $\boldsymbol{w}_{t} \sim \mathcal{N}(0, \Sigma_{\boldsymbol{w}})$ models uncertainty. The objective of RL algorithms is learning to accomplish a given task based on interaction data. The task is encoded in a cost function $c(\boldsymbol{x}_{t})$, defined to characterize the immediate penalty for being in state $\boldsymbol{x}_{t}$. The system inputs are chosen in accordance with a policy $\pi_{\boldsymbol{\theta}}: \boldsymbol{x} \mapsto \boldsymbol{u}$ that depends on the parameter vector $\boldsymbol{\theta}$. Then, the objective is to find the policy that minimizes the expected cumulative cost over a finite number of time steps $T$, with initial state distribution $p(\boldsymbol{x}_{0})$, i.e., $J(\boldsymbol{\theta}) = \sum_{t=0}^T \mathbb{E}_{\boldsymbol{x}_{t}}\left[c(\boldsymbol{x}_{t})\right]$.

\section{Method}\label{sec:proposed_approach}
MC-PILCO4PMS consists of the iteration of three phases: (i) model learning, (ii) policy optimization, and (iii) policy execution. In the first phase, MC-PILCO4PMS relies on GPR to estimate the one-step-ahead system dynamics, while for the optimization of the policy parameters, MC-PILCO4PMS implements a gradient-based strategy. In the following, we briefly discuss the two phases.

\subsection{Model Learning}\label{sec:modelLearning}
\textbf{Dynamics model.} 
The proposed one-step-ahead GP model exploits the intrinsic correlation between the position and velocity. In our algorithm a distinct GP model is learned to predict the velocity change, while positions are obtained by integration. This approach is different from previous GP-based MBRL algorithms, such as \cite{deisenroth2011pilco,parmas2018pipps,chatzilygeroudis2017black}, that learn one independent model for each state component.\\
Let us indicate the components of $\boldsymbol{q}_t$ and $\boldsymbol{\dot{q}}_t$ with $q_t^{(i)}$ and $\dot{q}_t^{(i)}$, respectively, where $i \in \{1,\ldots, d_{\boldsymbol{q}}\}$. Then, let $\Delta^{(i)}_t = \dot{q}^{(i)}_{t+1} - \dot{q}^{(i)}_{t}$ be the difference between the value of the \textit{i}-th velocity at time $t+1$ and $t$, and $y^{(i)}_{t}$ the noisy measurement of $\Delta^{(i)}_t$. For each velocity component $i$, we model $\Delta^{(i)}_t$ with a distinct GP with zero mean and kernel function $k(\cdot,\cdot)$, which takes as input $\tilde{\boldsymbol{x}}_{t} = [\boldsymbol{x}_{t}, \boldsymbol{u}_{t}]$. Details on the kernel choice can be found in Appendix~\ref{app:kernel}. 
In GPR the posterior distribution of $\Delta^{(i)}_t$ given the data is Gaussian, with mean and covariance available in closed form, see \cite{williams2006gaussian}. Then, given the GP input $\tilde{\boldsymbol{x}}_{t} = [\boldsymbol{x}_{t}, \boldsymbol{u}_{t}]$, a prediction of the velocity changes $\hat{\Delta}_t^{(i)}$ can be sampled from the aforementioned posterior distribution. When considering a sufficiently small sampling time $T_s$, it is reasonable to assume constant accelerations between two consecutive time-steps, and the predicted positions and velocities are obtained with the following equations, $\hat{q}_{t+1}^{(i)} = q_{t}^{(i)} +T_s \dot{q}_{t}^{(i)} +\frac{T_s}{2}\hat{\Delta}_t^{(i)} $ and $\hat{\dot{q}}_{t+1} = \dot{q}_{t}^{(i)} + \hat{\Delta}_t^{(i)}$ for $i \in \{1,\dots, d_{\boldsymbol{q}}\}$.

\textbf{Training data computation.}
As described before, velocities are not accessible and have to be estimated from measurements of positions. Notice that the velocity estimates used to train the GP models can be computed offline, exploiting the (past and future) history of measurements to improve accuracy. Well filtered data, that resemble the real states of the system, improve significantly the adherence between the learnt model and the real system. In our experiments, we computed offline the velocities used to train the GPs, using for example, the central difference formula, i.e., $\dot{\boldsymbol{q}}_t = (\boldsymbol{q}_{t+1}-\boldsymbol{q}_{t-1})/(2 T_s)$, which is an acausal filter.  We would like to underline that these state estimates are different from the ones computed real-time and provided to the control policy during system interaction. Typically, due to real-time constraints, online estimates are less accurate and it is fundamental to keep this in consideration during policy optimization as we can see in the following.


\subsection{Policy optimization}
MC-PILCO4PMS optimizes the policy parameters with a gradient-based strategy. At each optimization step the algorithm performs the following operations: (i) approximation of the cumulative cost relying on a Monte Carlo approximation; (ii) computation of the gradient and update of $\boldsymbol{\theta}$. More precisely, the algorithm samples $M$ particles from the initial state distribution $p(\boldsymbol{x}_0)$, and  simulates their evolution for $T$ steps. At each simulation step the inputs are selected in accordance with the current policy, and the next state is predicted with the GP models previously described. This procedure models the propagation of the model uncertainty for long-term predictions. Then, the Monte Carlo estimate of the cumulative cost is $\hat{J}(\boldsymbol{\theta}) = \sum_{t=0}^T \left( \frac{1}{M}\sum_{m=1}^M c\left(\boldsymbol{x}_t^{(m)}\right)\right)$, 
where $\boldsymbol{x}_t^{(m)}$ denotes the state of the \textit{m}-th particle at time $t$. The gradient is computed by backpropagation on the computational graph of $\hat{J}(\boldsymbol{\theta})$, exploiting the reparametrization trick \cite{kingma2013auto} to propagate the gradient through the stochastic operations, i.e., the sampling from the GP posteriors distribution. Advantages of MC based long-term predictions w.r.t to e.g., moment matching \cite{deisenroth2011pilco} are that no assumptions on the state distribution and on the kernel function in the GP models have to be made. The policy parameters are updated using the Adam solver \cite{kingma2014adam}. In the remainder of this section we describe the particles simulation, which is the main novelty introduced to deal with PMS.  

\begin{wrapfigure}{r}{0.5\linewidth}
  \begin{center}
    \includegraphics[width=0.48\textwidth]{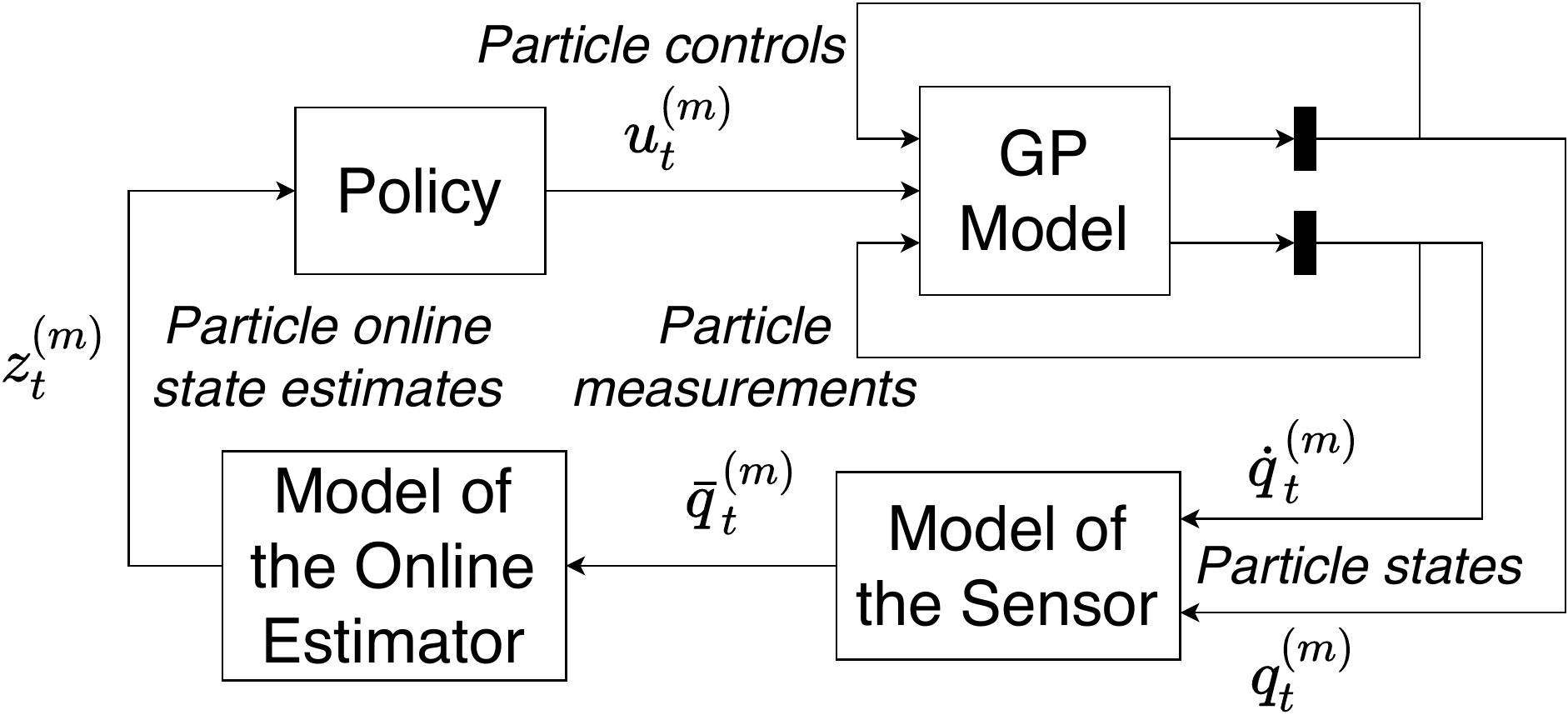}
  \end{center}
  \caption{\small Block schemes illustrating particles generation in MC-PILCO4PMS.}\label{fig:block}
\end{wrapfigure}

\textbf{Particles simulation with PMS.} In order to deal with PMS we not only simulate the evolution of the system state, but also the evolution of the observed states, modeling the measurement system and the online state observers implemented in the real system. A block scheme of the particles generation is reported in Fig.\ref{fig:block}. Let $\boldsymbol{x}^{(m)}_t=[\boldsymbol{q}^{(m)}_t,\dot{\boldsymbol{q}}^{(m)}_t]$ be the state of the \textit{m}-th particle at the simulation step $t$ predicted by the GP models. In order to transform the prediction of the system state to the observed state, firstly, we simulate the measurement system by corrupting positions with a zero mean Gaussian i.i.d noise $\boldsymbol{e}^{(m)}_t$: $\bar{\boldsymbol{q}}^{(m)}_t = \boldsymbol{q}^{(m)}_t + \boldsymbol{e}^{(m)}_t$. Secondly, the measured states are used to compute an estimate of the observed states: $\boldsymbol{z}^{(m)}_t = f_z(\bar{\boldsymbol{q}}^{(m)}_t\dots \bar{\boldsymbol{q}}^{(m)}_{t-m_q},\boldsymbol{z}^{(m)}_{t-1}\dots \boldsymbol{z}^{(m)}_{t-1-m_z})$, where $f_z$ is the online state observer implemented in the real system, with memory $m_q$ and $m_z$. Finally, the control inputs for each particle are computed as $\pi(\boldsymbol{z}^{(m)}_t)$, the next particles states are sampled from the GP dynamics, and the procedure is iterated. The procedure aims at obtaining robustness w.r.t. delays and distortions introduced by measurement noise and online observers. Notice that selecting the inputs as $\pi(\boldsymbol{x}^{(m)}_t)$, as done in several previous MBRL algorithms, is equivalent to assume full access to the system state,
which is often an unrealistic assumption when dealing with real systems, since the difference between the system state and the observed state might be significant. This is a key differentiation of our method. Let us denote, MC-PILCO, the version of the proposed algorithm which assumes fully access to the system state during policy optimization. A numerical comparison between MC-PILCO and two state-of-the-art GP-based MBRL algorithms is reported in the Appendix~\ref{app:comparison}. The results obtained show that MC-PILCO overperforms both the algorithms in terms of data-efficiency and accuracy.

\section{Experiments}\label{sec:experiments}
MC-PILCO4PMS has been tested both in simulation and in real systems. First, we validate in simulation the impact of taking into consideration the measurement system and the online filter during particle simulation. Second, MC-PILCO4PMS has been successfully applied to two real systems: a Furuta pendulum and a Ball-and-Plate system. Further details about the implementation of the algorithm on the presented systems can be found in Appendices \ref{app:cartpole}, \ref{app:furuta}, \ref{app:bp}.

\begin{figure}[!tb]
\minipage{0.32\textwidth}
\centering
  \includegraphics[width=1\linewidth]{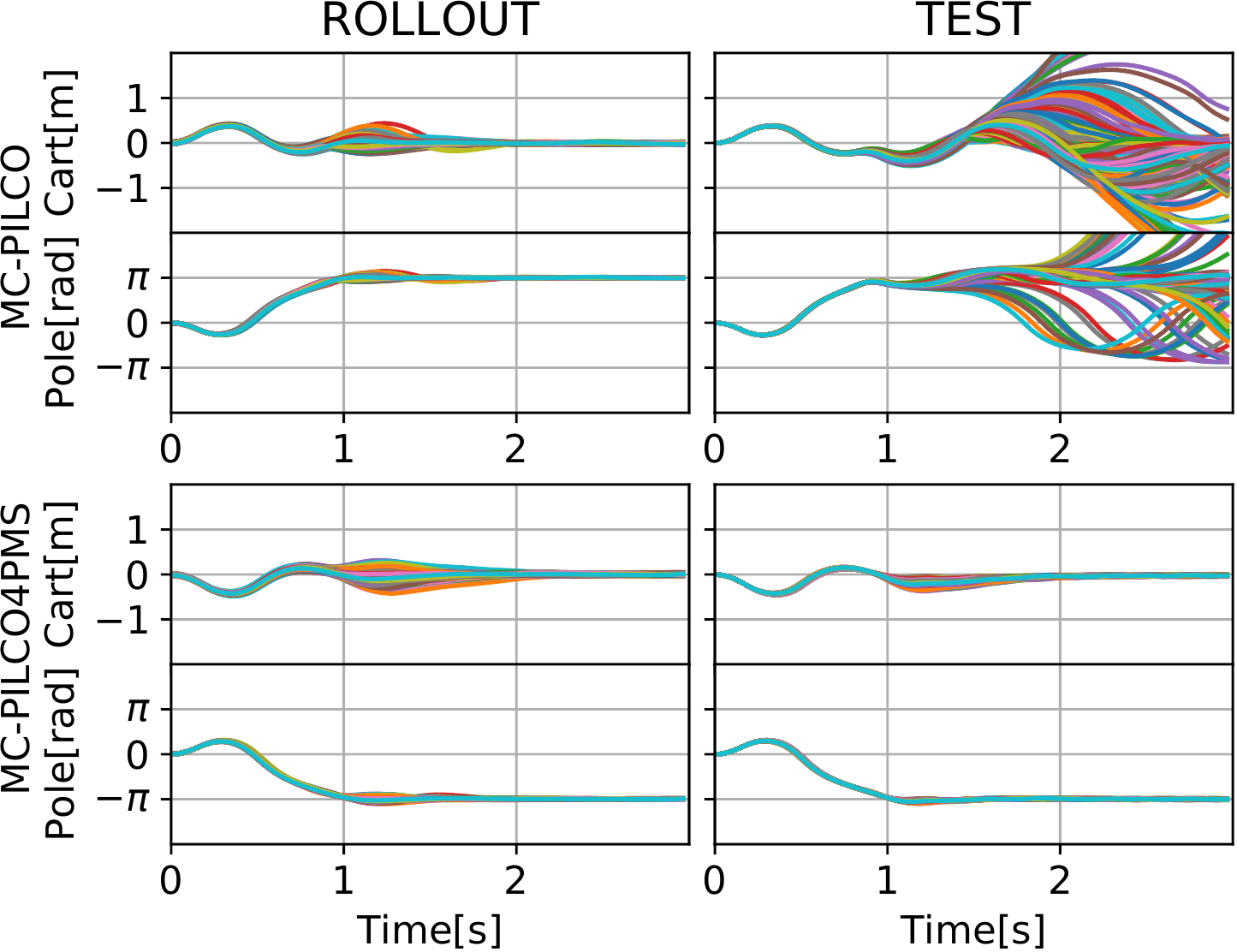}
  \caption{\small Comparison of 400 simulated particles rollout (left) and the trajectories performed applying repetitively the policy 400 times in the system (right) with the simulated cart-pole system. MC-PILCO results are on the top plots, while MC-PILCO4PMS are on the bottom.}
  \label{fig:filt_comparison}
\endminipage\hfill
\minipage{0.32\textwidth}
\centering
  \includegraphics[width=\linewidth]{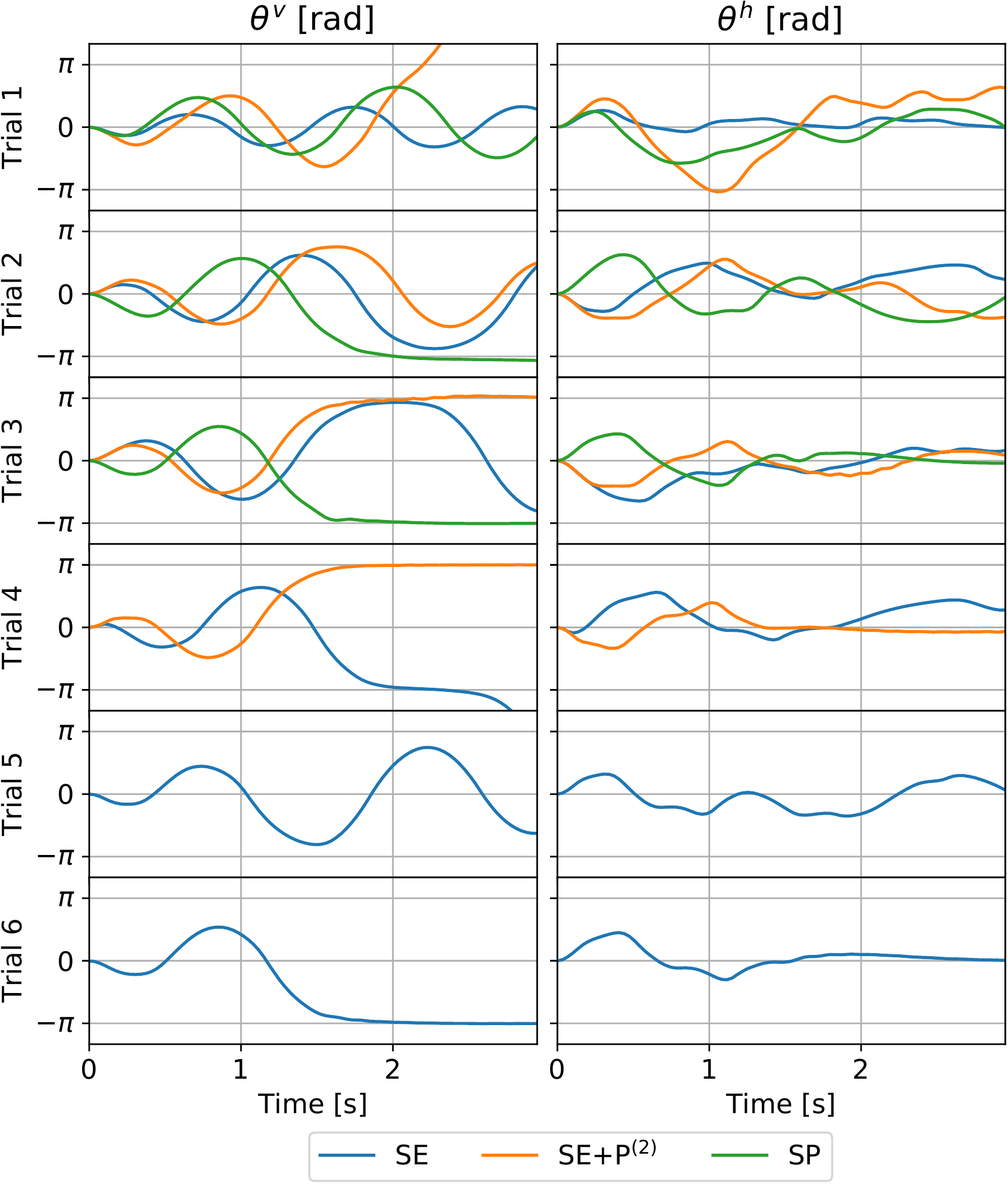}
  \caption{\small Trajectories for the pendulum angle (left) and arm angle (right) obtained at each trial. For all the kernels, the angles are plotted up to the trial that solved the task.}
  \label{fig:FP_results}
\endminipage\hfill
\minipage{0.32\textwidth}%
\centering
  \includegraphics[width=\linewidth]{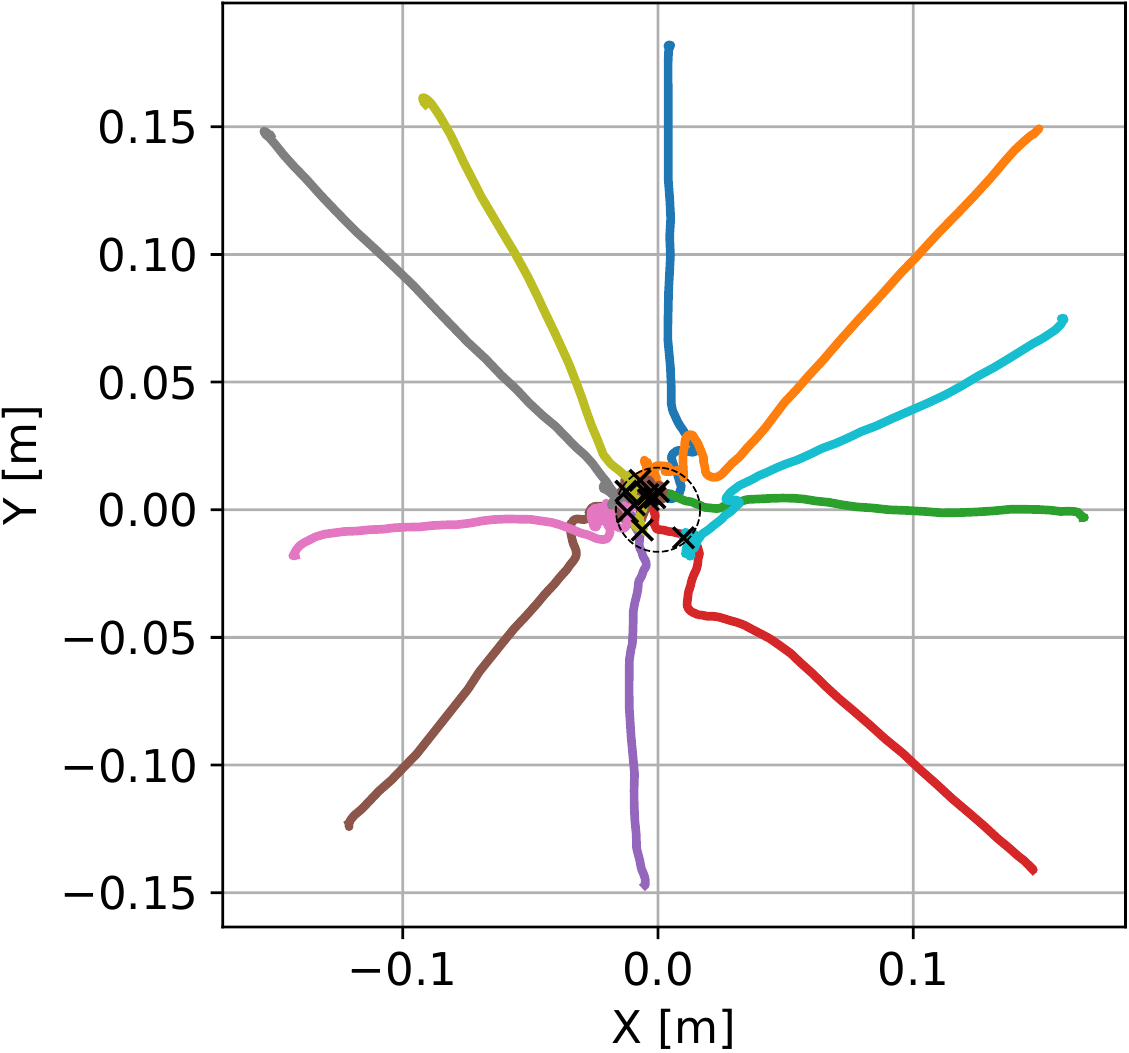}
  \caption{\small Ten different ball trajectories obtained on the Ball-and-Plate under the final policy learned by MC-PILCO4PMS. Steady-state positions are marked with black crosses. The dashed circle has the same diameter of the used ball.}
  \label{fig:b&p}
\endminipage
\end{figure}

\textbf{Simulation as proof of concepts.}
Here, we test the relevance of modeling the presence of online observers on a simulated cart-pole system. The objective is to learn a policy able to swing-up 
the pole and stabilize it in the upwards equilibrium, while keeping the cart stationary. We assumed to be able to measure only the cart position and the pole angle. The online estimates of the velocities were computed by means of causal numerical differentiation followed by a first order low-pass filter. The velocities used to train the GPs were derived with the central difference formula. Two policy functions have been trained: the first has been derived with MC-PILCO, assuming direct access to the full state predicted by the model; the second policy has been derived using MC-PILCO4PMS. In Figure \ref{fig:filt_comparison}, we report the results of a Monte Carlo study with 400 runs. Even though the two policies perform similarly when applied to the learned models, the results obtained with the cart-pole system are significantly different. MC-PILCO4PMS solves the task in all 400 attempts. In contrast, in several attempts, the MC-PILCO policy does not solve the task, due to delays and discrepancies introduced by the online filter and not considered during policy optimization.

\textbf{Furuta Pendulum.}
The Furuta pendulum \cite{cazzolato2011furuta} is a popular under-actuated benchmark system that consists of a driven arm, revolving in the horizontal plane, with a pendulum attached to its end, which rotates in the vertical plane. Let $\theta^h$ be the horizontal angle of the arm, and $\theta^v_t$ the vertical angle of the pendulum. The objective is to learn a controller able to swing-up the pendulum and stabilize it in the upwards equilibrium ($\theta_t^v=\pm\pi$ [rad]) with $\theta_t^h=0$ [rad]. Offline estimates of velocities for the GP model have been computed by means of central differences. Causal numerical differentiation were used for the online estimation. MC-PILCO4PMS managed to solve the task using the three different choices of kernel functions presented in Appendix \ref{app:kernel}. In Figure \ref{fig:FP_results}, we show the resulting trajectories for each trial. These experiments show the effectivness of MC-PILCO4PMS and confirm the higher data efficiency of more structured kernels, which is one of the advantage that MC-PILCO4PMS offers by allowing any kernel function while in methods like PILCO the kernel choice is limited. For best of our knowledge, with 9 [s] of training data this algorithm is the most data-efficient to solve a~FP.

\textbf{Ball-and-Plate.}
The ball-and-plate system is composed of a square plate that can tilt in two orthogonal directions by means of two motors. On top of it, there is a camera to track the ball and measure its position on the plate. 
The objective of the experiment is to learn how to control the motor angles in order to stabilize the ball around the center of the plate. Measurements provided by the camera are very noisy, and cannot be used directly to estimate velocities from positions. We used a Kalman smoother \cite{einicke2006optimal} for the offline filtering of ball positions and associated velocities. Instead,  in real-time we used a Kalman filter \cite{kalman1960new} to estimate online the ball state from noisy measures of positions. MC-PILCO4PMS learnt a policy to stabilize the ball around the center starting from any initial position after the third trial, 11.33 [s] of interaction with the system. We tested the learned policy starting from ten different points, see Figure \ref{fig:b&p}. The mean steady-state error, i.e. the average distance of the final ball position from the center observed in the ten trials, was 0.0099 [m], while the maximum measured error was 0.0149 [m], which is lower than the ball radius of 0.016 [m]. 

\section{Conclusions}
We have presented a MBRL algorithm called, MC-PILCO4PMS, which does not assume that all the components of the state can be measured and we successfully applied it to robotic systems. The algorithm employs GPs to derive a probabilistic model of the system dynamics. Policy parameters are updated through a Monte Carlo gradient-based strategy: expected cumulative cost is estimated by averaging over hundreds of simulated rollouts, and policy gradient is computed by backpropagation on the resulting computational graph. We showed the importance of manipulating the measurements to both provide accurate state estimates to the model learning algorithm and to reproduce the measurement system together with the online state observer during policy optimization. 

\bibliography{references}
\newpage 
\section{Appendix}

\subsection{Kernel functions}\label{app:kernel}
One of the advantages of the particle-based policy optimization method is the possibility of choosing any kernel functions without restrictions. Hence, we considered different kernel functions as examples to model the evolution of physical systems. But the reader can consider a custom kernel function appropriate for his application.

\textbf{Squared exponential (SE)}. The SE kernel represents the standard choice adopted in many different works. It is defined as 
\begin{equation}\label{eq:SE}
    k_{SE}(\tilde{\boldsymbol{x}}_{t_j},\tilde{\boldsymbol{x}}_{t_k}) := \lambda^2 e^{-||\tilde{\boldsymbol{x}}_{t_j}-\tilde{\boldsymbol{x}}_{t_k}||^2_{\Lambda^{-1}}}\text{,}
\end{equation}
where the scaling factor $\lambda$ and the matrix $\Lambda$ are kernel hyperparameters which can be estimated by marginal likelihood maximization. Typically, $\Lambda$ is assumed to be diagonal, with the diagonal elements named lengthscales.

\textbf{SE + Polynomial (SE+$\text{P}^{(d)}$)}. Recalling that the sum of kernels is still a kernel \cite{williams2006gaussian}, we considered also a kernel given by the sum of a SE and a polynomial kernel. In particular, we used the Multiplicative Polynomial (MP) kernel, which is a refinement of the standard polynomial kernel, introduced in \cite{libera2019novel}. The MP kernel of degree $d$ is defined as the product of $d$ linear kernels, namely, 
\begin{equation*}
    k_{P}^{(d)}(\tilde{\boldsymbol{x}}_{t_j},\tilde{\boldsymbol{x}}_{t_k}) := \prod_{r=1}^ d\left(\sigma^2_{P_r} + \tilde{\boldsymbol{x}}_{t_j}^T\Sigma_{P_r} \tilde{\boldsymbol{x}}_{t_k}\right)\text{.}
\end{equation*}
where the $\Sigma_{P_r}>0$ matrices are distinct diagonal matrices. The diagonal elements of the $\Sigma_{P_r}$, together with the $\sigma^2_{P_r}$ elements are the kernel hyperparameters. The resulting kernel is
\begin{equation}\label{eq:SE+Pkernel}
    k_{SE+P^{(d)}}(\tilde{\boldsymbol{x}}_{t_j},\tilde{\boldsymbol{x}}_{t_k}) =k_{SE}(\tilde{\boldsymbol{x}}_{t_j},\tilde{\boldsymbol{x}}_{t_k}) + k_{P}^{(d)}(\tilde{\boldsymbol{x}}_{t_j},\tilde{\boldsymbol{x}}_{t_k})\text{.}
\end{equation}
The idea motivating this choice is the following: the MP kernel allows capturing possible modes of the system that are polynomial functions in $\tilde{\boldsymbol{x}}$, which are typical in mechanical systems \cite{GIP}, while the SE kernel models more complex behaviors not captured by the polynomial kernel.

\textbf{Semi-Parametrical (SP)}. When prior knowledge about the system dynamics is available, for example given by physics first principles, the so called physically inspired (PI) kernel can be derived. The PI kernel is a linear kernel defined on suitable basis functions $\phi(\tilde{\boldsymbol{x}})$, see for instance \cite{romeres2019semiparametrical}. More precisely, $\boldsymbol{\phi}(\tilde{\boldsymbol{x}}) \in \mathbb{R}^{d_{\phi}}$ is a, possibly nonlinear, transformation of the GP input $\tilde{\boldsymbol{x}}$ determined by the physical model. Then we have
\begin{equation*}
    k_{PI}(\tilde{\boldsymbol{x}}_{t_j},\tilde{\boldsymbol{x}}_{t_k}) = \boldsymbol{\phi}^T(\tilde{\boldsymbol{x}}_{t_j}) \Sigma_{PI}\boldsymbol{\phi}(\tilde{\boldsymbol{x}}_{t_k}) \text{,}
\end{equation*}
where $\Sigma_{PI}$ is a $d_{\phi} \times d_{\phi}$ positive-definite matrix, whose elements are the $k_{PI}$ hyperparameters; to limit the number of hyperparameters, a standard choice consists in considering $\Sigma_{PI}$ to be diagonal. To compensate possible inaccuracies of the physical model, it is common to combine $k_{PI}$ with an SE kernel, obtaining so called semi-parametric kernels \cite{SP_peters, romeres2019semiparametrical}, expressed as
\begin{equation*}\label{eq:SP_kernel}
    k_{SP}(\tilde{\boldsymbol{x}}_{t_j},\tilde{\boldsymbol{x}}_{t_k}) = k_{PI}(\tilde{\boldsymbol{x}}_{t_j},\tilde{\boldsymbol{x}}_{t_k}) + k_{SE}(\tilde{\boldsymbol{x}}_{t_j},\tilde{\boldsymbol{x}}_{t_k}) \text{.}
\end{equation*}
The rationale behind this kernel is the following: $k_{PI}$ encodes the prior information given by the physics, and $k_{SE}$ compensates for the dynamical components unmodeled in $k_{PI}$.

\subsection{Simulated Cart-pole}\label{app:cartpole}
The physical properties of the cart-pole system considered are the following: the masses of both cart and pole are 0.5 [kg], the length of the pole is $L=0.5$ [m], and the coefficient of friction between cart and ground is 0.1. The state at each time step $t$ is defined as $\boldsymbol{x}_t=[p_t, \dot{p}_t, \theta_t, \dot{\theta}_t]$, where $p_t$ represents the position of the cart and $\theta_t$ the angle of the pole. The target states corresponding to the swing-up of the pendulum is given by $p^{des}=0$ [m] and $\vert \theta^{des} \vert = \pi$ [rad]. The downward stable equilibrium point is defined at $\theta_t = 0$ [rad]. As done in \cite{deisenroth2011pilco}, in order to avoid singularities due to the angles, $\boldsymbol{x}_t$ is replaced with the state representation $\bar{\boldsymbol{x}}_t = [p_t, \dot{p}_t, \dot{\theta}_t , sin(\theta_t), cos(\theta_t)]$ inside GP inputs. For the GP models SE kernels have been chosen \eqref{eq:SE}. The control action is the force that pushes the cart horizontally. We considered white measurement noise with standard deviation of $3\cdot 10^{-3}$, and as initial state distribution $\mathcal{N}([0,0,0,0],\text{diag}([10^{-4},10^{-4},10^{-4},10^{-4}]))$. In order to obtain reliable estimates of the velocities, samples were collected at 30 [Hz]. The number of particles has been set to $M=400$ in all the tests.

The cost function optimized in MC-PILCO is the following,
\begin{equation}\label{eq:abs_cost_cartpole}
    c(\boldsymbol{x}_t) = 1-\text{exp}\left(-\left(\frac{|\theta_t|-\pi}{l_{\theta}}\right)^2 -\left(\frac{p_t}{l_p}\right)^2\right),
\end{equation}
where $l_{\theta}$ and $l_p$ are named lengthscales. Notice that the lengthscales define the shape of $c(\cdot)$, the cost function goes to its maximum value more rapidly with small lengthscales. Therefore, higher cost is associated to the same distance from the target state with lower $l_{\theta}$ and $l_p$. The lower the lengthscale the more selective the cost function. The absolute value on $\theta_t$ is needed to allow different swing-up solutions to both the equivalent target angles of the pole $\pi$ and $-\pi$. The selected lengthscales were $l_{\theta}=3$ and $l_p=1$.

The policy adopted is an RBF network policy with outputs limited by an hyperbolic tangent function, properly scaled. We call this function \textit{squashed-RBF-network} and it is defined as
\begin{equation}\label{eq:policy}
    \pi_{\boldsymbol{\theta}}(\bar{\boldsymbol{x}}_t) = u_{max}\;\text{tanh} \left(\frac{1}{u_{max}}\sum_{i=1}^{n_b} w_i e^{||\boldsymbol{a}_i-\bar{\boldsymbol{x}}_t||_{\Sigma_{\pi}}^2}\right)\text{.}
\end{equation}
parameters are $\boldsymbol{\theta} = \left\{\boldsymbol{w}, A,\Sigma_{\pi}\right\}$, where $\boldsymbol{w}=[w_1\dots w_{n_b}]$ and $A=\left\{\boldsymbol{a}_1\dots\boldsymbol{a}_{n_b}\right\}$ are, respectively, the weights and the centers of the $n_b$ Gaussian basis functions, while ${\Sigma_{\pi}}$ determines their shapes; in all experiments we assumed ${\Sigma_{\pi}}$ to be diagonal. $u_{max}$ is the maximum control action applicable. For this experiment we choose $n_b=200$ basis functions and $u_{max}=10$ [N]. The exploration trajectory is obtained by applying at each time step a random control action sampled from $\mathcal{U}(-u_{max},u_{max})$.

\subsection{Comparison with state of the art algorithms}\label{app:comparison}
\begin{table}[t]
    \centering
    \begin{tabular}{|l|l|l|l|l|l|}
  \hline
  & \small{Trial 1} & \small{Trial 2} & \small{Trial 3} & \small{Trial 4} & \small{Trial 5} \\ \hline
  \small{PILCO} & \small{2\%} & \small{4\%} & \small{20\%} & \small{36\%} & \small{42\%} \\ \hline
  \small{Black-DROPS} & \small{0\%} & \small{4\%} & \small{30\%} & \small{68\%} & \small{86\%} \\ \hline
  \small{MC-PILCO} & \small{0\%} & \small{14\%} & \small{78\%} & \small{94\%} & \small{100\%} \\ \hline
 \end{tabular}
    \caption{\small Success rate per trial obtained with PILCO, Black-DROPS and MC-PILCO.}
    \label{tab:success_rates}
\end{table}
We tested PILCO\cite{deisenroth2011pilco}, Black-DROPS\cite{chatzilygeroudis2017black} and MC-PILCO on the simulated cart-pole system. \begin{wrapfigure}{r}{0.5\linewidth}
\centering
  \includegraphics[width=\linewidth]{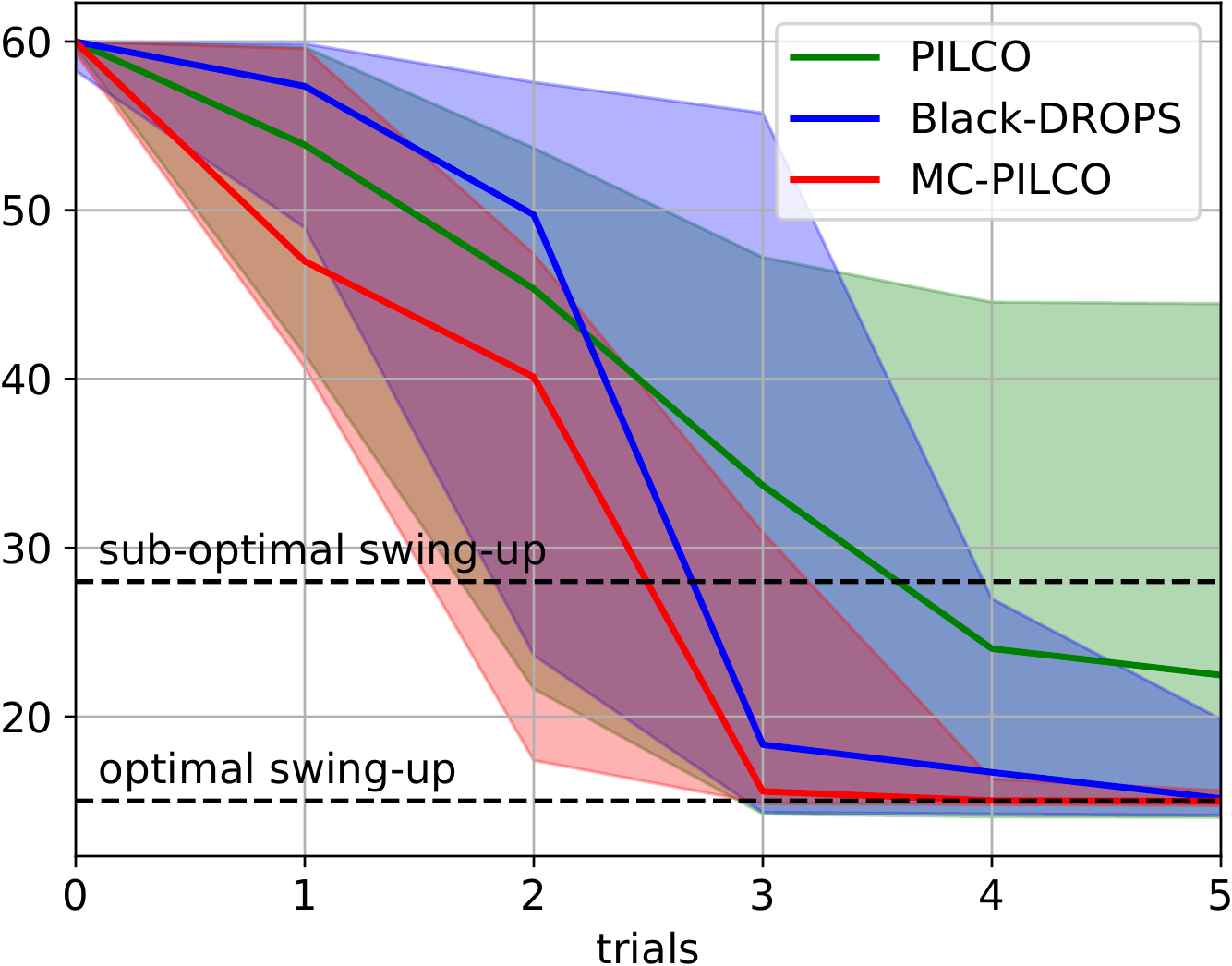}
  \caption{\small Median and confidence interval of the cumulative cost $c^{pilco}(\cdot)$ per trial obtained with PILCO, Black-DROPS and MC-PILCO.}
  \label{fig:confronto_algoritmi} 
\end{wrapfigure}
The setup is equal to the one described in Appendix~\ref{app:cartpole}, with the only two difference that here the samples are collected at 20 [Hz] and the noise standard deviation is $10^{-2}$. In PILCO and Black-DROPS, we considered their original cost function, \begin{equation}\label{eq:pilco_cost}
c^{pilco}(\boldsymbol{x}_t) = 1- \text{exp}\left(-\frac{1}{2} \left(\frac{d_t}{0.25}\right)^2 \right),
\end{equation}
where $d_t^2=p_t^2 + 2 p_t L sin(\theta_t) + 2 L^2 (1+ cos(\theta_t))$ is the squared distance between the tip of the pole and its position at the unstable equilibrium point with $p_t=0$ [m]. This last cost is also adopted as a common metric to compare the results obtained by the three algorithms. Results of the cumulative cost are reported in Figure \ref{fig:confronto_algoritmi}, observed success rates are shown in Table \ref{tab:success_rates}. MC-PILCO achieved the best performance both in transitory and at convergence, by trial 5, it learned how to swing up the cart-pole with a success rate of 100\%. In each and every trial, MC-PILCO obtained cumulative costs with lower median and less variability. On the other hand, the policy in PILCO showed poor convergence properties with only 42\% of success rate after all the 5 trials. Black-DROPS outperforms PILCO, but it obtained worse results than MC-PILCO in each and every trial, with a success rate of only 86\% at trial 5.

\subsection{Furuta Pendulum}\label{app:furuta}
The Furuta pendulum (FP) \cite{cazzolato2011furuta} is a popular benchmark system used in nonlinear control and reinforcement learning. The system is composed of two revolute joints and three links. \begin{wrapfigure}{r}{0.35\linewidth}
  \begin{center}
    \includegraphics[width=0.30\textwidth]{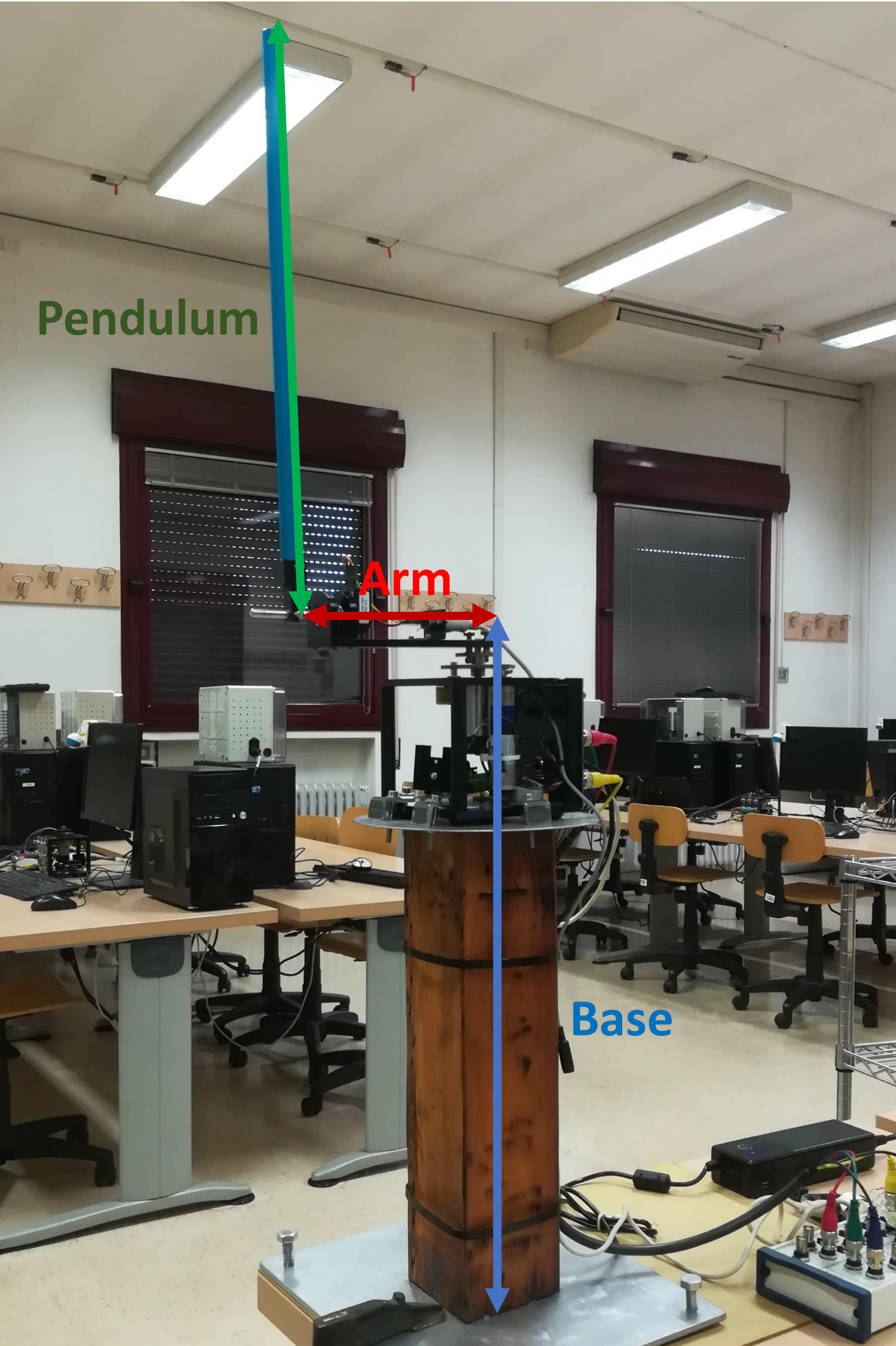}
  \end{center}
  \caption{\small Furuta pendulum used in the experiment while being controlled in the upward equilibrium point by the learned policy.}\label{fig:fp_links}
\end{wrapfigure}  
The first link, called the base link, is fixed and perpendicular to the ground. The second link, called arm, rotates parallel to the ground, while the rotation axis of the last link, the pendulum, is parallel to the principal axis of the second link, see Figure~\ref{fig:fp_links}.
The FP is an under-actuated system as only the first joint is actuated. In particular, in the FP considered the horizontal joint is actuated by a DC servomotor, and the two angles are measured by optical encoders with  4096 [ppr]. The control variable is the motor voltage.
Let the state at time step $t$ be $\boldsymbol{x}_t = [\theta^h_t, \dot{\theta}^h_t, \theta^v_t, \dot{\theta}^v_t ]^T$, where $\theta^h_t$ is the angle of the horizontal joint and $\theta^v_t$ the angle of the vertical joint attached to the pendulum. The objective is to learn a controller able to swing-up the pendulum and stabilize it in the upwards equilibrium ($\theta_t^v=\pm\pi$ [rad]) with $\theta_t^h=0$ [rad]. The trial length is 3 [s] with a sampling frequency of 30 [Hz]. The cost function is defined as
\begin{equation}\label{eq:furuta cost}
    c(\boldsymbol{x}_t)=1-\text{exp}\left( -\left(\frac{\theta_t^h}{2}\right)^2  -\left(\frac{|\theta_t^v|-\pi}{2}\right)^2 \right) +c_{b}(\boldsymbol{x}_t),
\end{equation}
with
\begin{align*}
    c_{b}(\boldsymbol{x}_t) =& \frac{1}{1+\text{exp}\left(-10\left(-\frac{3}{4}\pi-\theta^h_t \right)\right)}\\
    &+\frac{1}{1+\text{exp}\left(-10\left(\theta^h_t-\frac{3}{4}\pi \right)\right)}\text{.}
\end{align*}

The first part of the function in \eqref{eq:furuta cost} aims at driving the two angles towards $\theta_t^h=0$ and $\theta_t^v=\pm\pi$, while $c_{b}(\boldsymbol{x}_t)$ penalizes solutions where $\theta_t^h \leq -\frac{3}{4} \pi$ or  $\theta_t^h \geq \frac{3}{4} \pi$. We set those boundaries to avoid the risk of damaging the system if the horizontal joint rotates too much. Offline estimates of velocities for the GP model have been computed by means of central differences. For the online estimation, we used causal numerical differentiation: $\dot{\boldsymbol{q}}_t = (\boldsymbol{q}_{t}-\boldsymbol{q}_{t-1})/T_s$, where $T_s$ is the sampling time. Instead of $\boldsymbol{x}_t$, we considered the extended state $\bar{\boldsymbol{x}}_t=[\dot{\theta}^h_t, \dot{\theta}^v_t, sin(\theta^h_t),cos(\theta^h_t),sin(\theta^v_t),cos(\theta^v_t)]^T$ in GP input. The policy is a \textit{squashed-RBF-network} with $n_b=200$ basis functions that receives as input $\bar{\boldsymbol{z}}_t=[(\theta^h_t-\theta^h_{t-1})/{T_s}, (\theta^v_t-\theta^v_{t-1})/T_s, sin(\theta^h_t),cos(\theta^h_t),sin(\theta^v_t),cos(\theta^v_t)]^T$. We used 400 particles to estimate the policy gradient from model predictions. The exploration trajectory has been obtained using as input a sum of ten sine waves of random frequencies and same amplitudes. The initial state distribution is assumed to be $\mathcal{N}([0,0,0,0]^T,\text{diag}([5 \cdot 10^{-3},5 \cdot 10^{-3},5 \cdot 10^{-3},5 \cdot 10^{-3}])$.  $M=400$ particles were used for gradient estimation.

\subsection{Ball-and-Plate}\label{app:bp}
The ball-and-plate system is composed of a square plate that can be tilted in two orthogonal directions by means of two motors. On top of it, there is a camera to track the ball and measure its position on the plate. Let $(b^x_t,b^y_t)$ be the position of the center of the ball along X-axis and Y-axis, while $\theta^{(1)}_t$ and $\theta^{(2)}_t$ are the angles of the two motors tilting the plate, at time $t$. So, the state of the system is defined as $\boldsymbol{x}_t = [b^x_t,b^y_t,\dot{b}^x_t,\dot{b}^y_t,\theta^{(1)}_t,\theta^{(2)}_t,\dot{\theta}^{(1)}_t,\dot{\theta}^{(2)}_t]^T$. The drivers of the motors allow only position control, and do not provide feedback about the motors angles. To keep track of the motor angles, we defined the control actions as the difference between two consecutive reference values sent to the motor controllers, and we limited the maximum input to a sufficiently small value, such that the motor controllers are able to reach the target angle within the sampling time. Then, in first approximation, the reference angles and the motor angles coincide, and we have $u_t^{(1)} = \theta^{(1)}_{t+1}-\theta^{(1)}_t$ and $u_t^{(2)} = \theta^{(2)}_{t+1}-\theta^{(2)}_t$. The objective of the experiment is to learn how to control the motor angles in order to stabilize the ball around the center of the plate. Notice that the control task, with the given definition of inputs, is particularly difficult because the policy must learn to act in advance, and not only react to changes in the ball position.

The cost function is defined as
\begin{equation*}
    c(\boldsymbol{x}_t)=1-\text{exp}\left(-g_t(\boldsymbol{x}_t) \right), \qquad \text{with}
\end{equation*}
\begin{equation*}
    g_t(\boldsymbol{x}_t) = \left(\frac{b^x_t}{0.15}\right)^2 +\left(\frac{b^y_t}{0.15}\right)^2  +\left(\theta_t^{(1)}\right)^2  +\left(\theta_t^{(2)}\right)^2.
\end{equation*}

The trial length is 3 [s], with a sampling frequency of 30 [Hz]. Measurements provided by the camera are very noisy, and cannot be used directly to estimate velocities from positions. We used a Kalman smoother for the offline filtering of ball positions $b^x_t,b^y_t$ and associated velocities $\dot{b}^x_t,\dot{b}^y_t$. In the control loop, instead, we used a Kalman filter \cite{kalman1960new} to estimate online the ball state from noisy measures of positions. Concerning the model, we need to learn only two GPs predicting the evolution of the ball velocity because we directly control motor angles, hence, their evolution is assumed deterministic. GP inputs, $\tilde{\boldsymbol{x}}_t = [\bar{\boldsymbol{x}}_t, u_t]$, include an extended version of the state,  $\bar{\boldsymbol{x}}_t=[b^x_t,b^y_t,\dot{b}^x_t,\dot{b}^y_t,sin(\theta^{(1)}_t),cos(\theta^{(1)}_t),sin(\theta^{(2)}_t),cos(\theta^{(2)}_t),(\theta^{(1)}_t-\theta^{(1)}_{t-1})/T_s,(\theta^{(2)}_t-\theta^{(2)}_{t-1})/T_s]^T$ where angles have been replaced by their sines and cosines, and motor angular velocities have been estimated with causal numerical differentiation ($T_s$ is the sampling time). 
\begin{wrapfigure}{r}{0.35\linewidth}
  \begin{center}
    \includegraphics[width=\linewidth]{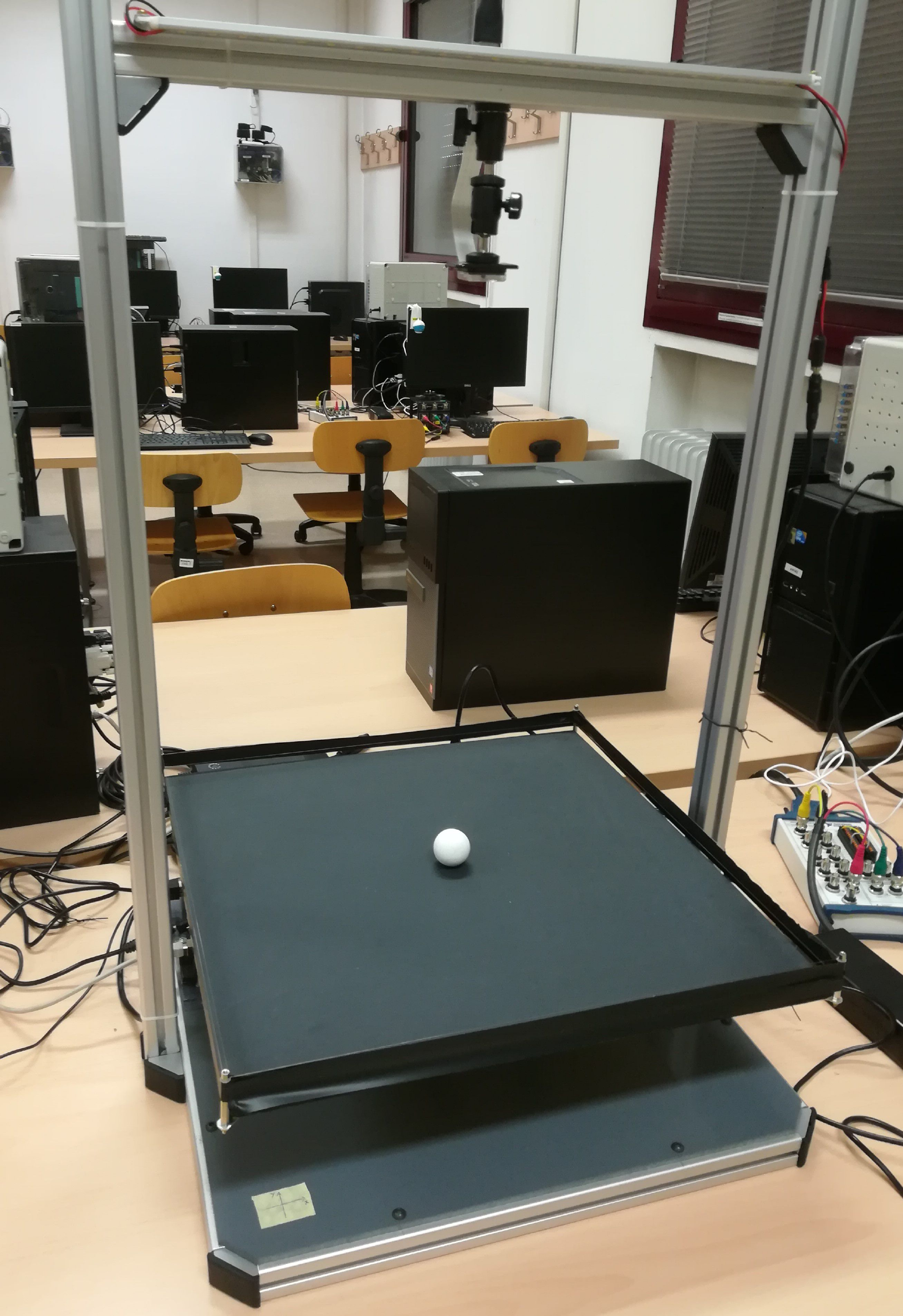}
  \end{center}
  \caption{\small Ball-and-plate system used in the experiment.}\label{fig:bp_picture}
\end{wrapfigure}
The SE+$\text{P}^{(1)}$ kernel \eqref{eq:SE+Pkernel} is used, where the linear kernel acts only on a subset of the model inputs, $\tilde{\boldsymbol{x}}^{lin}_t=[sin(\theta^{(1)}_t), sin(\theta^{(2)}_t), cos( \theta^{(1)}_t), cos(\theta^{(2)}_t), u_t]$. We considered $M=400$ particles for policy gradient estimation. The policy is a multi-output \textit{squashed-RBF-network}, with $n_b=400$ basis functions, that receives as inputs the estimates of  $(b^x_t,b^y_t,\dot{b}^x_t,\dot{b}^y_t,\theta^{(1)}_t, \theta^{(1)}_{t-1},\theta^{(2)}_t,\theta^{(2)}_{t-1})$ computed with the Kalman filter; maximum angle displacement is $u_{max}=4$ [deg] for both motors. Initial exploration is given by two different trials, in which the control signals are two triangular waves perturbed by white noise. Mostly during exploration and initial trials, the ball might touch the borders of the plate. In those cases, we kept data up to the collision instant. A peculiarity of this experiment in comparison to the others seen before is a wide range of initial conditions. In fact, the ball could be positioned anywhere on the plate's surface, and the policy must control it to the center. The initial distribution of $b^x_0$ and $b^y_0$ is a uniform $\mathcal{U}(-0.15,0.15)$, which covers almost the entire surface (the plate is a square with sides of about 0.20 [m]). For the other state components, $\theta^{(1)}_t$ and $\theta^{(2)}_t$, we assumed tighter initial distributions $\mathcal{U}(-10^{-6},10^{-6})$.

\end{document}